\documentclass[10pt,twocolumn,letterpaper]{article}

\usepackage{cvpr}      % To produce the REVIEW version
\usepackage[dvipsnames]{xcolor}

\definecolor{cvprblue}{rgb}{0.21,0.49,0.74}
\usepackage[pagebackref,breaklinks,colorlinks,citecolor=cvprblue]{hyperref}
\usepackage{amssymb}
\usepackage{colortbl} 
\usepackage{makecell}
\usepackage{mathtools, nccmath}
\usepackage{tabularx,arydshln}
\usepackage[accsupp]{axessibility}
\DeclarePairedDelimiter{\nint}\lfloor\rceil

\usepackage{bbm}
\usepackage{amsthm}
\usepackage{fontawesome}
\usepackage{bbding}
\usepackage{pifont}
\newcommand{\cmark}{\ding{51}}
\newcommand{\xmark}{\ding{55}}

\newtheorem{assumption}{Assumption}[section]
\newtheorem{definition}{Definition}[section]
\theoremstyle{remark}

\usepackage[ruled,linesnumbered]{algorithm2e}

\usepackage{accents}
\usepackage{multirow}
\makeatletter
\def\widebar{\accentset{{\cc@style\underline{\mskip18mu}}}}
\makeatother

\newcommand{\improvedperf}[1]{({\color{blue}{\footnotesize \textbf{+#1\%}}})}
\newcommand{\decreasedperf}[1]{({\color{gray}{\footnotesize \textbf{-#1\%}}})}
\title{Retraining-free Model Quantization via One-Shot Weight-Coupling Learning}

\author{
Chen Tang$^{1*}$ \quad Yuan Meng$^{1*}$ \quad Jiacheng Jiang$^{1}$ \quad Shuzhao Xie$^{1}$ \quad Rongwei Lu$^{1}$  \\ Xinzhu Ma$^{2}$ \quad Zhi Wang$^{1 \dagger}$ \quad Wenwu Zhu$^{1 \dagger}$ \\
{
\normalsize
$^{1}$Tsinghua University \quad $^{2}$The Chinese University of Hong Kong 
}\\ 
}

\begin{document}
\maketitle

{\def\thefootnote{*}\footnotetext{\scriptsize Equal contributions. \quad $^{\dagger}$Corresponding authors. \quad Z. Wang is also with TBSI. }}

\begin{abstract} 
Quantization is of significance for compressing the over-parameterized deep neural models and deploying them on resource-limited devices.  
Fixed-precision quantization suffers from
performance drop due to the limited numerical representation ability. 
Conversely, mixed-precision quantization (MPQ) is advocated to compress the model effectively by allocating heterogeneous bit-width for layers. MPQ is typically organized into a searching-retraining two-stage process. Previous works only focus on determining the optimal bit-width configuration in the first stage efficiently, while ignoring the considerable time costs in the second stage and thus hindering deployment efficiency significantly. 
In this paper, we devise a one-shot training-searching paradigm for mixed-precision model compression. 
Specifically, in the first stage, all potential bit-width configurations are coupled and thus optimized simultaneously within a set of shared weights.
However, our observations reveal a previously unseen and severe bit-width interference phenomenon among highly coupled weights during optimization, leading to considerable performance degradation under a high compression ratio.
To tackle this problem, we first design a bit-width scheduler to dynamically freeze the most turbulent bit-width of layers during training, to ensure the rest bit-widths converged properly. 
Then, taking inspiration from information theory, we present an information distortion mitigation technique to align the behaviour of the bad-performing bit-widths to the well-performing ones. 
In the second stage, an inference-only greedy search scheme is devised to evaluate the goodness of configurations without introducing any additional training costs. 
Extensive experiments on three representative models and three datasets demonstrate the effectiveness of the proposed method. 
Code can be available on \href{https://www.github.com/1hunters/retraining-free-quantization}{https://github.com/1hunters/retraining-free-quantization}. 
\end{abstract} 
\vspace{-0.7cm} 
\section{Introduction}
\label{sec:intro}
Recent years have witnessed the tremendous achievements made by deep network-driven applications, ranging from classification \cite{he2016deep,howard2017mobilenets,sandler2018mobilenetv2}, object detection \cite{tan2020efficientdet,ren2015faster,redmon2018yolov3}, and segmentation \cite{chen2017deeplab,maskrcnn}. 
However, the huge number of parameters in these deep models poses intractable challenges for both training \cite{chen2022bert2bert,lu2023dagc,huang2019gpipe} and inference \cite{polino2018model,han2015deep}. 
To enable efficient deep learning on inference, several techniques are proposed, including pruning \cite{molchanov2019importance,liu2018rethinking}, quantization \cite{tang2022mixed,esser2020learned,zhou2016dorefa}, and neural architecture search \cite{howard2019searching,tang2023elasticvit}. 

Ultra-low bit-width neural network quantization \cite{zhou2016dorefa,esser2020learned,uhlich2019mixed} is an appealing model compression technique to simplify the hardware complexity and improve the runtime efficiency of over-parameterized deep models. However, under severely limited numerical representation capabilities, performing such compression across the whole neural network usually incurs an unacceptable performance drop. 
Mixed-precision quantization (MPQ) \cite{wang2019haq,habi2020hmq,cai2020rethinking,huang2022sdq,tang2022mixed}, by allocating unequally bit-width for weight and activation tensors of each layer, can largely avoid accuracy degradation while maintaining the proper model size and runtime overhead (\eg on-device latency). 
The underlying principle of MPQ is that layers contribute very differently to the final accuracy \cite{cai2020rethinking,wang2019haq,tang2022mixed}, so the compression algorithm should apply heterogeneous precision rather than a uniform one across the whole network. 
Besides, hardware starts to support mixed-precision computation \cite{wang2019haq,chen2021towards} in these years, which further pushes the study for mixed-precision compression. 

Recently, MPQ has been extensively studied from several perspectives, \eg through reinforcement learning \cite{wang2019haq,elthakeb2020releq}, differentiable methods \cite{wang2021generalizable,cai2020rethinking}, and proxy-based approaches \cite{dong2019hawq,chen2021towards}. 
They all try to solve one challenge, that says, how to find the optimal bit-width configuration for each layer, in an exponentially large $\mathcal{O}(n^{2L})$ space, where $n$ is the number of bit-width candidates and $L$ is the number of layers in the deep network. 
To this end, they organize a \emph{searching-then-retraining pipeline}, in which the first stage aims to finish the bit-width allocation as fast as possible, and naturally becomes the focus of the research. Nevertheless, previous works tend to ignore the importance of the second stage, which in fact consumes considerable time cost for retraining the model to fit the obtained bit-width configurations (a.k.a, the policy). 
To recover the performance, LIMPQ \cite{tang2022mixed} needs about 200 GPU-hours to retrain a single ResNet18 policy. This impedes the real-world quantized mixed-precision model deployment---we might not have much time to retrain every policies for all devices. 

Instead, we consider a new paradigm termed as \emph{training-then-searching}, repositioning the resource-intensive training process to the forefront of the mixed-precision quantization pipeline. 
The initial stage focuses on the development of a weight-sharing quantization model, where all possible bit-width configurations are concurrently optimized within unified network of weights to fulfill extensive search requirements. 
Importantly, this weight-sharing model undergoes a singular training session and, notably, \emph{requires no subsequent retraining following the search}. 
Subsequently, in the second stage, we present an inference-only search employing a bidirectional greedy scheme to judiciously determine the optimal bit-width for each layer. 

\begin{figure*}[t]
\begin{center}
\includegraphics[width=1.0\textwidth]{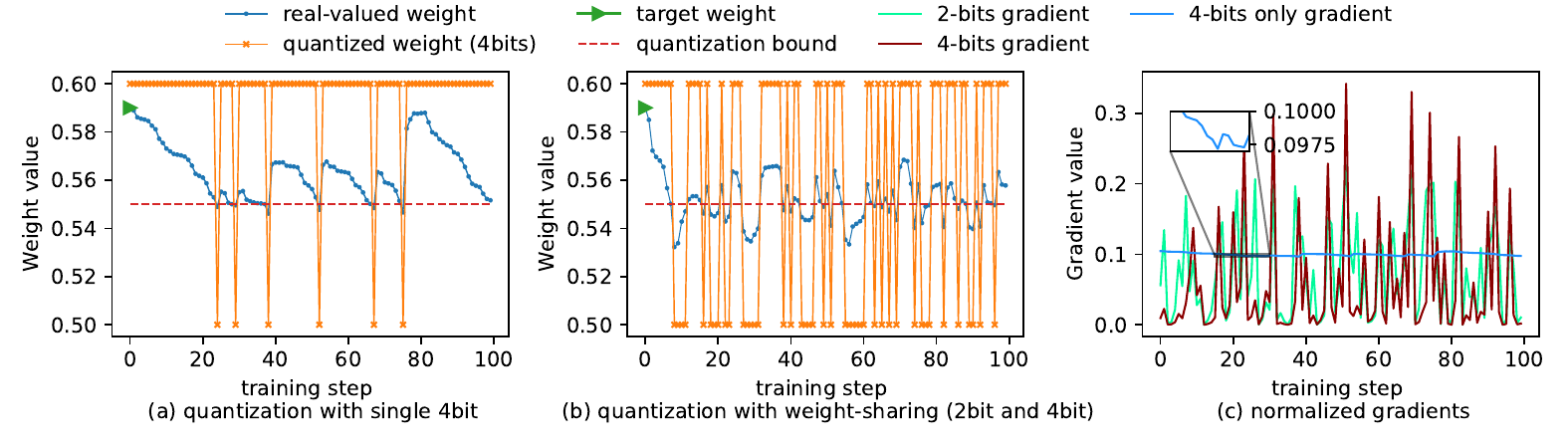}
\caption{\textbf{(a)} 2D regression of 
 \emph{single} 4-bits quantization, \textbf{(b)} 2D regression of 4-bits quantization with an additional 2-bits (\emph{i.e.,} weight-sharing quantization), and \textbf{(c)} the L2-normalized gradients of these two regressions. 
Compared with \cref{fig:2d_regression}(a), the weight in \cref{fig:2d_regression}(b) is more unstable due to the bit-width interference. Notably, the gradient of 4-bits also has a larger variance under weight-sharing. 
}
\label{fig:2d_regression} 
\end{center}
\vspace{-0.5cm}
\end{figure*}

The primary focus of this paper lies in the training of a high-quality weight-sharing quantization model, which highly relies on ingenious weight-coupling learning method with heterogeneous bit-widths. 
We identify a distinctive phenomenon inherent in weight-sharing quantization—referred to as the \textit{bit-width interference problem}. This problem arises from the highly shared weights between bit-widths, the same weight could be quantized to very different discrete values for various bit-widths, so significantly superimposed quantization noise of various bit-widths leads to daunting optimization challenges, as we will discuss later. 
We illustrate the bit-width interference problem in \cref{fig:2d_regression}, one can see that even the introduction of a single additional bit-width can cause the shared weight to frequently traverse quantization bound, resulting in training instability and large gradient variance. 

To understand and circumvent the issue of weight-sharing quantization, we conduct a detailed analysis of the bit-width interference problem (Sec.~\ref{sec:interference}). Building upon this understanding, we introduce a bit-width scheduler designed to freeze the bit-widths that contribute to weight interference, ensuring proper convergence for the remaining bit-widths. Furthermore, during dynamic training, we observe an information distortion phenomenon associated with the unfrozen bit-widths. To mitigate this distortion, we propose to align the behavior of poorly performing bit-widths with their well-performing counterparts. 
Extensive experiments demonstrate that these two complementary techniques not only unravel the intricacies of the bit-width interference problem but also provide meaningful performance improvements of weight-sharing quantization models. 
To summarize, our contributions are as follows: 
\begin{itemize} 
    \item We identify and analyze the bit-width interference problem in weight-sharing quantization models, revealing its impact on optimization challenges, training stability, and convergence. 
    \item To train the weight-sharing quantization model, we first design a novel bit-width scheduler that freezes interfering bit-widths during training, ensuring proper convergence and addressing instability caused by the introduction of additional bit-widths. 
    \item We also propose a strategy inspired by information theory to align poorly performing bit-widths with their well-performing counterparts, mitigating information distortion during dynamic training and enhancing the overall performance. 
    \item Extensive experiments on three representative models and three benchmarks demonstrate the effectiveness of proposed method.
    For example, under an average 4-bit constraint, our method leads on ResNet with a top accuracy of 71.0\% at only 31.6G BitOPs and no retraining cost, compared to the second-best at 70.8\% accuracy with higher 33.7G operations and 90 epochs of retraining. 
    
\end{itemize}

\begin{figure}[t]
\includegraphics[width=0.45\textwidth]{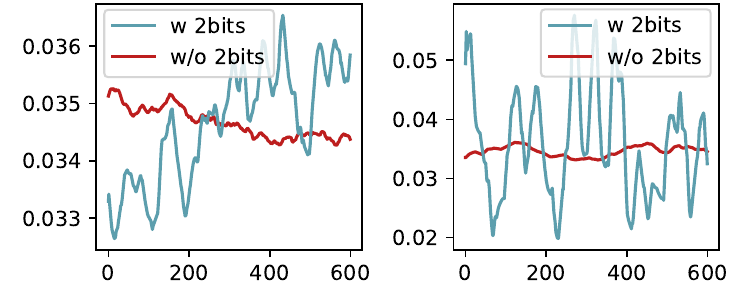}
\caption{Distance between full-precision latent weights and quantized weights on MobileNetV2 of a point-wise conv layer. Left: 4-bits. Right: 6-bits. } 
\label{fig:distance_mobilenetv2}
\end{figure}
\section{Related Work}
\label{sec:related_work}
\subsection{Neural Network Quantization} 
In this paper, we only consider the context in quantization-aware training, as it can achieve higher compression ratio than post-training quantization \cite{nagel2020up,hubara2021accurate}. 
Quantization can be generally categorized into two classes: fixed-precision quantization and mixed-precision quantization. \\
\textbf{Fixed-Precision Quantization.} 
Fixed-precision quantization involves assigning a uniform bit-width to all layers. Specifically, methods such as Dorefa \cite{zhou2016dorefa} and PACT \cite{choi2018pact} employ a low-precision representation for weights and activations during forward propagation. They leverage the Straight-Through Estimation (STE) technique \cite{bengio2013estimating} to estimate the gradient of the piece-wise quantization function for backward propagation. 
LSQ \cite{esser2020learned} scales the weight and activation distributions by introducing learnable step-size scale factors for quantization functions. 
\\ 
\textbf{Mixed-Precision Quantization.} 
Mixed-Precision Quantization (MPQ) delves into the intricate aspects of low-precision quantization by recognizing the inherent variability in redundancy across different layers of the deep model. 
By allocating smaller bit-widths to layers with high redundancy, MPQ optimizes model complexity without causing a significant performance decline. The challenge, however, lies in determining the most suitable bit-width for each layer, considering that the bit-width selection is a discrete process, and the potential combinations of bit-width and layer (referred to as ``policy'') grow exponentially. 

Strategies such as HAQ \cite{wang2019haq} and ReleQ \cite{elthakeb2020releq} leverage reinforcement learning (RL) techniques to derive a bit-width allocator. 
SPOS \cite{guo2020single}, EdMIPS \cite{cai2020rethinking}, BP-NAS \cite{yu2020search}, GMPQ \cite{wang2021generalizable} and SEAM \cite{tang2023seam} adopt differential neural architecture search (NAS) methods to learn the bit-width. However, these methods require to both search-from-scratch and train-from-scratch for the models when changing the search constraints. 
Unlike learning the optimal MPQ policy, HAWQ \cite{dong2019hawq,dong2020hawq} and MPQCO \cite{chen2021towards} use the Hessian information as the quantization sensitivity metrics to assist bit-width assignment. 
LIMPQ \cite{tang2022mixed} proposes to learn the layer-wise importance within a single quantization-aware training cycle. 

\subsection{Deep Learning with Weight-Sharing} 
Weight-sharing is an effective and practical technique to reuse weight to deal with joint task learning or avoid the need to store multiple copies of weights. 
Generally speaking, there have been two relevant topics to weight-sharing with this work, covering neural architecture search (NAS) and dynamic neural network. \\
\textbf{Neural Architecture Search.} 
NAS~\cite{howard2019searching,tan2019efficientnet,zoph2016neural} aims to automatically discover the well-performing topology of deep neural network in a vast search space, which composes of various operators (\eg convolutional layers with different kernel-size or channels). 
To improve the search efficiency, recent works \cite{pham2018efficient,guo2020single,tang2023elasticvit} both adopt the idea of weight-sharing to stuff the candidates into a shared and large meta-topology (a.k.a. the super-network). 
Weight-sharing allows to train directly the meta-topology and derive a sub-topology from the meta-topology to evaluate. 
This significantly shortens the evaluation time of the goodness for a given topology \cite{liu2018darts,pham2018efficient}. 
Although certain MPQ research \cite{cai2020rethinking,wang2021generalizable} leverages this NAS-style searching process, they still pay significant time for retraining. 
\\ 

\noindent
\textbf{Dynamic Neural Network.} 
Unlike conventional neural networks which are architecturally fixed during inference, dynamic neural networks \cite{han2021dynamic} enable dynamic computational paths on demand according to various input samples or running environments. 
For example, runtime channel pruning \cite{lin2017runtime,wang2020dynamic} dynamically activates channels of layers and layer skipping  \cite{wang2018skipnet,wang2022skipbert,shen2020fractional} adjusts depths of layers for different input images. 
To support the adaptive inference, the weight-sharing idea is applied to avoid substantial copies of weights. 
Therefore, they can both achieve a better accuracy-efficiency trade-off compared to their static counterparts. 
These have also been several works of dynamic bit-width neural networks \cite{jin2020adabits,tang2022arbitrary,bulat2021bit,xu2023eq}. 
However, they either provide only the fixed-precision quantization that supports very limited bit-width configurations or have to drop the ultra-low bit-widths (\eg 2 bits, 3bits, \etc) to guarantee the training convergence but lose the chance for achieving high compression ratio. 
\section{Methodology}
\subsection{Preliminary} 
\textbf{Quantization.} 
The uniform quantization function under $b$ bits in quantization-aware training (QAT) 
maps the input full-precision activations and weights to the homologous quantized values $[0, 2^{b}-1]$ and $[-2^{b-1}, 2^{b-1}-1]$. The quantization functions $Q_b(\cdot)$ that quantize the input values $x$ to quantized values $\hat{x}$ can be expressed as follows:

\begin{equation} 
\hat{x}=Q_b(x;\gamma)= \nint{\text{\rm clip}(\frac{x}{\gamma},{\rm N_{min}}, {\rm N_{max}})} \times \gamma_b, \frac{\partial \nint{x}}{\partial x} \triangleq 1, 
\label{eq_quantization_preliminary}
\end{equation}
where $\nint{\cdot}$ is the rounding-to-nearest function, and
$\gamma$ is the scale factor. 
% \mxz{what is the x? is the input x or v?} 
The Straight-Through Estimation (STE) is used to pass the gradients for back-propagation. 
The ${\rm clip}$ function ensures that the input values fall into the range $[{\rm N_{min}}, {\rm N_{max}}]$ \cite{esser2020learned,zhou2016dorefa}. 
For ${\rm ReLU}$ activations, ${\rm N_{min}} = 0$ and ${\rm N_{max}} = 2^{b}-1$. For weights, ${\rm N_{min}} =-2^{b-1}$ and ${\rm N_{max}} = 2^{b-1}-1$. 
$\gamma_b$ is the learnable scalar parameter used to adjust the quantization mappings, called the \emph{step-size scale factor}. 
For a network, each layer has two distinct scale factors in the weights and activations quantizer. \\
\textbf{Weight-Sharing for Mixed-Precision Quantization.} 
We consider a model $f(\cdot) = f_{\rm \textbf W_{L-1}} \circ ... \circ f_{\rm \textbf W_0} (\cdot) $ with $L$ layers, and each layer has $N$ bit-width choices. 
Let ${\rm \textbf W:= \{{\rm \textbf W}_l\}}_{l=0}^{L-1}$ be the set of flattened weight tensors of these $L$ layers. 
Therefore, the corresponding search space $\mathcal{A}$ with $N^{2L}$ mixed-precision quantization policies $\{(b^{(w)}_l, b^{(a))}_l\}_{l=0}^{L-1}$ share the same \textbf{latent weights} $\rm \textbf W$. 
To track the time-prohibitive training costs of traversing the whole search space, Monte-Carlo sampling is used to approximate the expectation term \cite{tang2022arbitrary,yu2020bignas,tang2022mixed}. 
The overall optimization objective is formulated as follows
\begin{equation} 
\begin{aligned}
 & \mathop{\arg\min}_{\rm \textbf W} \quad \mathbb{E}_{\mathcal{S} \thicksim \mathcal{A}}\left[\mathcal{L}(f(\textbf{x}; \mathcal{S}, w^{(\mathcal{S})}), \textbf{y})\right] \\
& \approx \mathop{\arg\min}_{\rm \textbf W} \frac{1}{K}\sum_{\mathcal{S}_k \thicksim \text{\rm U}(\mathcal{A})}^K\left[\mathcal{L}(f(\textbf{x}; \mathcal{S}_k, \hat{\rm \textbf W}^{(\mathcal{S}_k)}), \textbf{y})\right], 
\end{aligned}
\label{eq:weight_sharing}
\end{equation}
where $\hat{\rm \textbf W}^{(\mathcal{S}_k)}$ is the quantized weights of $k$-th sampled policy $\mathcal{S}_k$ derived from the latent weights $\rm \textbf W$. This weight-sharing of layer $l$ is achieved by 
\begin{equation}
    \small 
    \hat{\rm \textbf W}^{(\mathcal{S}_k)} := \{\hat{{\rm \textbf W}}^{(\mathcal{S}_k)}_l\}_{l=0}^{L-1}, \quad \text{where} \quad
    \hat{\rm \textbf W}_{l}^{(\mathcal{S}_k)} = Q_{b^{(k)}_{l, {\rm w}} \in \mathcal{S}_k}({\rm \textbf W}_{l};\gamma) 
\end{equation}
according to \cref{eq_quantization_preliminary}, where $b^{(k)}_{l, {\rm w}} \in \mathcal{S}_k$ is the bit-width of weight of $l$-th layer in the policy $\mathcal{S}_k$. 

\begin{table}[t]
    \centering
    \caption{Accuracy of the weight-sharing quantization with/without low bit-width for MobileNetv2 (@80 epochs). }
    \fontsize{8.7}{13.8}\selectfont
    \begin{tabular}{ccc}
    \toprule
        ~ & \makecell{Top-1 Acc. (\%) \textbf{\emph{w/}} 2bits \\ { ( $\downarrow$ sampling probability)}}  & Top-1 Acc. (\%) \textbf{\emph{w/o}} 2bits \\ \midrule
        6 bit  & 69.2 & 70.4 \\
        4 bit  & 68.1 & 69.1 \\
        % 6 bit  & -  & - & - & - \\
        % 4 bit  & -  & - & - & - \\
    \bottomrule
    \end{tabular}
    
    \label{tab:bit-width_interference}
\end{table}

\subsection{Interference in Highly Coupled Weight-sharing} 
\label{sec:interference}
While training is possible, there still many challenges exist in weight-sharing in practice. 
For example, ABN \cite{tang2022arbitrary} observes 
 a large accuracy gap between the lower bit-widths and higher bit-widths. 
These works simply bypass this problem and remove the ultra-low bit-width until the training becomes stable, however, doing so does not achieve high compression ratios. 

Here, we discuss the bit-width interference caused by highly coupled latent weights $\rm \textbf W$. Suppose we have $K=2$ sampled policies \{$\mathcal{S}_0$,  $\mathcal{S}_{1}$\} at training step $t$ in \cref{eq:weight_sharing}, and the bit-width of weight is sampled differently, namely $\hat{\rm \textbf W}^{(\mathcal{S}_0)} \neq \hat{\rm \textbf W}^{(\mathcal{S}_{1})}$. 
$\forall b^{(0)}_{l, {\rm w}} \in \mathcal{S}_{0}, \forall b^{(1)}_{l, {\rm w}} \in \mathcal{S}_{1} : b^{(0)}_{l, {\rm w}} < b^{(1)}_{l, {\rm w}}
$. 
\begin{assumption}[Non-uniform Bit-width Convergence]
Quantized weights $\hat{\rm \textbf W}_{l}^{(\mathcal{S}_1)} = Q_{b^{(1)}_{l, {\rm w}} \in \mathcal{S}_1}({\rm \textbf W}_{l};\gamma)$ of bit-width $b_k$ at step $t$ is nearly converged while the $\hat{\rm \textbf W}_{l}^{(\mathcal{S}_0)} = Q_{b^{(0)}_{l, {\rm w}} \in \mathcal{S}_0}({\rm \textbf W}_{l};\gamma)$ is not converged properly. The smaller and not fully converged bit-width in $\mathcal{S}_0$ will pose negative impact to the larger but well converged bit-width in $\mathcal{S}_1$. 
\label{asm:training_convergence} 
\end{assumption}
The situation in Assump.~\ref{asm:training_convergence} is observed in weight-sharing network when the learning capacity gap between sub-networks is large enough \cite{yu2020bignas,tang2022arbitrary}. 
To further analyze the impact of $\mathcal{S}_0$, we can approximates the loss perturbation with the second-order Taylor expansion: 
\begin{footnotesize}
\begin{equation} 
\begin{aligned}
     \Delta \mathcal{L} &= \sum_{n=1}^N \ell ( f(\hat{\rm \textbf W}^{(\mathcal{S}_1)} + \Delta {\rm \textbf W}, \text{x}^{n}), \text{y}^{n})
     - \sum_{n=1}^N \ell ( f(\hat{\rm \textbf W}^{(\mathcal{S}_1)}, \text{x}^{n}), \text{y}^{n})\\ 
    & \approx \nabla_{\hat{\rm \textbf W}^{(\mathcal{S}_1)}} \ell (\hat{\rm \textbf W}^{(\mathcal{S}_1)}) \Delta {\rm \textbf W} + \Delta {\rm \textbf W}^{\mathsf{T}} \nabla_{\hat{\rm \textbf W}^{(\mathcal{S}_1)}}^2 \ell (\hat{\rm \textbf W}^{(\mathcal{S}_1)}) \Delta {\rm \textbf W}, 
\end{aligned} 
\label{eq_assumption}
\end{equation}
\end{footnotesize}
where $\ell (\cdot)$ is the cross-entropy loss function, $\Delta {\rm \textbf W} := \{ {\Delta} {\rm \textbf W}^{(\mathcal{S}_1)}_l \}_{l=0}^{L-1}$ is the quantization noise introduced by policy $\mathcal{S}_1$ to policy $\mathcal{S}_0$ on each layer. 
It is noteworthy that the lower the bit-width, the larger the quantization noise introduced \cite{zhou2018adaptive,jin2020adabits,huang2022sdq}, caused by the large rounding and clipping error under very limited available discrete values, \eg quantization error for 2 bits is roughly 5$\times$ for 3 bits. 
Therefore, putting small bit-width (\eg 2bit) into the weight-sharing will lead to large loss perturbation $\Delta \mathcal{L}$ and disturb the overall performance eventually (see \cref{tab:bit-width_interference}). 
Accordingly, one can draw such conclusions in \cref{eq_assumption}: $\emph{(i)}$ 
removing the low bit-width is the simplest way to erase the effects of quantization noise but loses the chance to compress the model with high compression ratio and
$\emph{(ii)}$ one can track the loss perturbation through $\Delta {\rm \textbf W}$ to iteratively freeze the most unstable bit-width, which motivates our method described later in \cref{sec_dynamic_bits_schedule}.

To illustrate the bit-width interference in weight-sharing quantization, we use a 2D regression quantization experiment depicted in \cref{fig:2d_regression}. Our optimization objective minimizes the empirical risk \cite{nagel2022overcoming,defossez2022differentiable}: 
\begin{equation}
    \mathop{\arg\min}_w \mathbbm{E}_{ \textbf{x} \sim \mathcal{N}(0,1)} \Big[ \Vert \textbf{x} w^{*} - \textbf{x} Q_b(w, \gamma) \Vert_2^2 \Big], 
\end{equation}
where $w^{*}$ represents the target weight and $\textbf{x}$ is sampled from a normal distribution. In \cref{fig:2d_regression}(a), the single-bit optimization showcases relatively stable quantized weights, occasionally crossing the quantization boundary due to gradient approximation of STE \cite{bengio2013estimating}. Comparatively, weight-sharing quantization exhibits more instability and the \emph{weight moves closer to the quantization bound more frequently} (\cref{fig:2d_regression}(b)), even with higher variance in gradients as in \cref{fig:2d_regression}(c). 

This interference extends beyond toy regression to modern neural networks, shown in \cref{fig:distance_mobilenetv2} and \cref{tab:bit-width_interference}. 
Furthermore, \cref{fig:distance_mobilenetv2} demonstrates the distance between quantized weights of 6 and 4 bits in one training epoch. Removing the smallest bit-width (2 bits) notably stabilizes the higher bit-width curve. However, introducing extra small bit-widths induces significant random oscillations, signifying heightened model training instability. 

\subsection{Dynamic Bit-width Schedule} 
\label{sec_dynamic_bits_schedule}
\cref{eq_assumption} reveals the decomposition of overall quantization noise $\Delta {\rm \textbf W}$ into layer-specific perturbation components, offering a metric to identify unstable layers. 
Therefore, \emph{dynamically freezing the bit-width causing weight interference ensures proper convergence for remaining bit-widths during training}. However, direct use of \cref{eq_assumption} poses computational challenges, particularly in calculating the Hessian and quantization noise terms, prompting us to devise an alternative method. 

We approximate layer perturbations by focusing on rounding errors due to their significant impact on overall performance  \cite{han2021improving,nagel2022overcoming}. 
Rounding errors portray the distance between full-precision weights and their discrete quantization levels, and reach maximums when at the midpoint between two quantization levels (\ie, the \emph{quantization bound} in \cref{fig:2d_regression}) because the possible quantization levels change. 
In other words, the closer to the quantization bounds, the more unstable the weights are, and therefore the unstable weights are more vulnerable to the weight-sharing. 
Therefore, tracking the round errors provide effective proxies for constructing our bit-width scheduler. 
For clarity, we first definite the Bit-width Representation Set (BRS) as follows: 
\begin{definition}[Bit-width Representation Set] 
For bit-width $b$ under uniform weight quantization, the bit-width representation set $\Phi_b := \gamma \times \{ -2^{b-1}, ..., 0, ..., 2^{b-1}-1 \}$, representing $2^{b}$ decomposed values of discrete quantization levels according to \cref{eq_quantization_preliminary}. 
\label{def:quantization_levels} 
\end{definition} 
The midpoints between two adjacent elements in a BRS are quantization bounds, where they have a uniform distance $\gamma$. 
Given a pre-defined weight bit-width candidates $B^{(w)}$, we can accumulate bit-specific unstable weights for BRS of each bit-width of each layer's shared weights ${\rm \textbf {W}}_l$. 
Therefore, we calculate the unstable weight criterion $\hat{\Delta} \rm \textbf W^{\text{unstable}}$ by 
\begin{equation}
\begin{aligned}
     \hat{\Delta} {\rm \textbf W^{\text{unstable}}} & \triangleq \{\hat{\Delta} {\rm \textbf W^{\text{unstable}}_l }\}_{l=0}^{L-1},  \text{where}
    \\ 
      \hat{\Delta} {\rm \textbf W}_l^{\text{unstable}}  = & \sum_{b \in B^{(w)}} \quad \frac{1}{2^{b}} \frac{1}{\Vert {\rm \textbf {W}}_l \Vert_0} \cdot \\ 
     & \sum_{q_b \in \Phi_b} 
    \sum_{{\rm \textbf w}_{l, *} \in {\rm \textbf W}_l} \mathbbm{1}_{ \mid {\rm \textbf w}_{l, *} \mid \leq \gamma \times \left(\frac{1 - \epsilon}{2} + \frac{q_b}{\gamma} \right)}, 
\end{aligned} 
\label{eq:unstable_weight_criterion}
\end{equation} 
where $\epsilon \in [0, 1]$ is to control the range of weights we care about. 
After that, we choose the frozen layer set $\Omega$ with a Top-$\mathcal{K}$ selector from the weight criterion, 
\begin{equation}
\large
    \Omega \leftarrow \texttt{\textbf{TopKToFreeze}} (\hat{\Delta} {\rm \textbf W^{\text{unstable}}}; \mathcal{K}),
\end{equation}
and the smallest bit-width of selected layers in $\Omega$ will be temporarily frozen periodically. In practice, we use a cosine scheduler to gradually decrease the value of $\mathcal{K}$ to guarantee that more unstable low bits will be frozen early to improve the convergence of more high bits. 

\begin{figure}[t]
\includegraphics[width=0.47\textwidth]{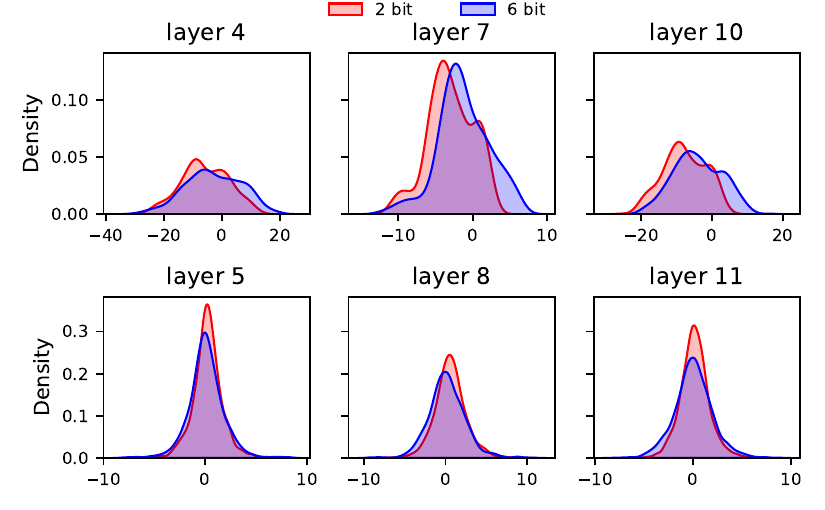}
\caption{Output density at 2bit and 6bits. Small bit-width shows noteworthy information distortion. } 
\label{fig:density_of_2_6bits}
\end{figure}

\subsection{Optimization during Dynamic Training} 
\textbf{Information Distortion Mitigation.} 
While freezing the bit-width of layers, we observe the outputs of the remaining small bit-widths of layers still exhibit a \textit{information distortion} compared to their high precision counterparts, as shown in \cref{fig:density_of_2_6bits}. 
Inspired by the information bottleneck principle \cite{tishby2015deep,xu2023q,tishby2000information}, we expect if the smallest bit-width is sampled of a layer $l$, its outputs ${\rm \textbf O}_{l}^{S}$ can preserve the information of its large counterparts ${\rm \textbf O}_{l}^{H}$. 
However, directly optimizing this mutual information term $I({\rm \textbf O}_{l}^{S}; {\rm \textbf O}_{l}^{H})$ is infeasible, so we consider a feature alignment loss function to optimize their rectified Euclidean distance as follows: 
\begin{equation}
\begin{aligned}
\mathbb{E} \Big[ \| & \max \{Q, {\small \frac{{\rm \textbf O}^{S} - \mu ({{\rm \textbf O}}^{S})}{\sqrt{\sigma ({{\rm \textbf O}}^{S}) + \zeta}}} \eta_{{\rm \textbf O}^{S}} + \xi_{{\rm \textbf O}^{S}} \} -\\   
& \max \{Q, {\small \frac{{\rm \textbf O}^{H} - \mu ({{\rm \textbf O}}^{H})}{\sqrt{\sigma ({{\rm \textbf O}}^{H}) + \zeta}}} \eta_{{\rm \textbf O}^{H}} + \xi_{{\rm \textbf O}^{H}} \}  \| \Big], 
    \label{eq:rectified_euclidean_distance}
\end{aligned}
\end{equation} 
where $\zeta$ is a small constant to avoid Divide-by-Zero errors, $\eta$ and $\xi$ are the learnable parameters for adapting the features, $\mu(\cdot)$ and $\sigma(\cdot)$ return the channel-wise mean and variance of input. 
\cref{eq:rectified_euclidean_distance} not only scales the features for better optimization but uses a $\max$ operator to avoid needless activations. See \cref{fig:density_of_2_6bits_ours} for visualization with proposed Information Distortion Mitigation technique. 
\\ 
\textbf{Fairness Weight Regularization.} 
Low-bit weights are actually subsets of high-bit weights, when a layer is sampled with different bit-width, low-bit weights will receive additional updates from high-bit weights. 
In other words, low-bit weights are subjected to very aggressive weight regularization, which exacerbates their underfitting issues \cite{yu2020bignas}. 
To ensure regularization fairness, we disable weight regularization for \emph{weights of current smallest bit-width} during training. 

\subsection{Bidirectional Greedy Search} 
To find the optimal quantization policy $\mathcal{S^*}$, the existing MPQ methods can be formulated to a bi-level optimization problem \cite{tang2022mixed}. 
In this paper, our well-trained weight-sharing model can serve as a good performance indicator to perform inference-only searching \cite{yu2020bignas,wang2020attentive}. 
This simplified procedure motivates us to devise a bidirectional greedy search scheme to determine the per-layer bit-width efficiently. 

Consider a mixed-precision quantization policy, $\mathcal{S}^{(t)}$, implemented at step $t$, with $L$ being the total number of layers. To evolve this policy to $\mathcal{S}^{(t+1)}$, rather than concurrently adjusting the bit-width of most layers (\eg, employing reinforcement learning), a step-by-step approach is taken. Specifically, the bit-width of a single layer is adjusted at a time, either increasing or decreasing by a single bit-width to create a provisional policy, $\mathcal{S}^{(t)}_i$, where $i \in \{0,...,2L-1\}$. This method yields a search space of complexity $\mathcal{O}(2L)$ for each iteration.
During each iteration, the permanent policy $\mathcal{S}^{(t+1)}$ is chosen in a greedy manner between these $2L$ policies, considering the trade-off between accuracy and efficiency, denoted as $J_i = \bar{\mathcal{L}}_{val} (\hat{\rm \textbf W}^{(\mathcal{S}_{i}^{(t)})}) + \lambda * \widebar{\texttt{\textbf{BitOps}}} (\mathcal{S}_{i}^{(t)})$ for each layer: 
\begin{equation} 
\begin{aligned}
    \mathcal{S}^{(t+1)} & \leftarrow \mathop{\arg\min}_i \left[ \Theta \right], \\
    \Theta \triangleq \{J_i | J_i = \bar{\mathcal{L}}_{val} & (\hat{\rm \textbf W}^{(\mathcal{S}_{i}^{(t)})}) + \lambda * \widebar{\texttt{\textbf{BitOps}}} (\mathcal{S}_{i}^{(t)}) \}_{i=0}^{2L-1}, 
\end{aligned} 
\end{equation} 
where $\bar{\mathcal{L}}$ and $\widebar{\texttt{\textbf{BitOps}}}$ are the min-max normalization loss and BitOPs to ensure their values fall into the interval $[0, 1]$, and $\lambda$ is the hyper-parameter to control the trade-off, respectively. By this means, the solution $\mathcal{S^*}$ is reached when the BitOPs is satisfied at final step $T$, \ie, $\mathcal{S}^{*} \leftarrow \mathcal{S}^{(T)}, \text{if} \  \texttt{\textbf{BitOps}}(\mathcal{S}^{(T)}) \leq C$. 
\begin{table}[t!]
\caption{Accuracy and efficiency results for ResNet. 
``Top-1 Acc.'' represents the Top-1 accuracy of the quantized model and full-precision model. 
``MP'' means mixed-precision quantization. 
``Retrain Cost'' denotes the epochs required to retrain the MPQ policy.  
``*'': reproduces through the vanilla ResNet architecture \cite{he2016deep}. 
The best results are bolded in each metric, the second best results are underlined. 
} 
\setlength{\tabcolsep}{1.0mm}
\centering
\fontsize{9.5}{13.8}\selectfont
\begin{tabular}{ccccc}
\toprule
Method                              &  \makecell{Bit-width \\(W/A)}  & \makecell{Top-1 Acc.\\(\%) $\uparrow$}   & \makecell{BitOPs \\(G) $\downarrow$} & \makecell{Retrain Cost \\(Epoch) $\downarrow$} \\ \midrule
Baseline                            & 32 / 32          &  70.5  &  -   & - \\
\hdashline
PACT \cite{choi2018pact}            & 3 / 3               & 68.1        & 23.0      & -        \\
LSQ$^*$  \cite{esser2020learned}                           & 3 / 3               & 69.4        & 23.0      & 90        \\
EWGS \cite{lee2021network} &  3 / 3        & 69.7 &  23.0    & 100 \\
EdMIPS \cite{esser2020learned}      & 3$_{\rm MP}$ / 3$_{\rm MP}$             & 68.2        & -          & 40      \\
GMPQ$^*$ \cite{wang2021generalizable}   & 3$_{\rm MP}$ / 3$_{\rm MP}$             & 68.6        & 22.8       & 40      \\
DNAS  \cite{wu2018mixed}     & 3$_{\rm MP}$ / 3$_{\rm MP}$             & 68.7        & 24.3      & 120        \\
FracBits  \cite{yang2021fracbits}  & 3$_{\rm MP}$ / 3$_{\rm MP}$             & 69.4        & 22.9      & 150    \\
LIMPQ  \cite{tang2022mixed}    & 3$_{\rm MP}$ / 3$_{\rm MP}$             & 69.7        & 23.0      & 90       \\
SEAM     \cite{tang2023seam}                & 3$_{\rm MP}$ / 3$_{\rm MP}$             & 70.0        & 23.0      & 90 \\
\rowcolor{yellow!25}Ours    & 2$_{\rm MP}$ / 3$_{\rm MP}$           &  67.7  & 17.3 & \textbf{0} \\
\rowcolor{yellow!25}Ours    & 3$_{\rm MP}$ / 3$_{\rm MP}$           & \textbf{70.2}   & 23.3 & \textbf{0} \\
\hdashline
PACT     \cite{choi2018pact}    
& 4 / 4               & 69.2        & 35.0      & - \\
LSQ$^*$  \cite{esser2020learned}            & 4 / 4               & 70.5        & 35.0      &  90 \\
EWGS \cite{lee2021network} &  4 / 4        & 70.6 &  35.0    & 100 \\
MPQCO    \cite{chen2021towards}             & 4$_{\rm MP}$ / 4$_{\rm MP}$             & 69.8      & -      & 30 \\
DNAS     \cite{gong2019differentiable}      & 4$_{\rm MP}$ / 4$_{\rm MP}$             & 70.6        & 35.1          & 120  \\
FracBits \cite{yang2021fracbits}            & 4$_{\rm MP}$ / 4$_{\rm MP}$             & 70.6        & 34.7       & 150    \\
LIMPQ    \cite{tang2022mixed}               & 4$_{\rm MP}$ / 4$_{\rm MP}$             & 70.8        & 35.0      & 90 \\
SEAM     \cite{tang2023seam}                & 4$_{\rm MP}$ / 4$_{\rm MP}$             & 70.8        & \underline{33.7}      & 90 \\
\rowcolor{yellow!25}Ours    & 4$_{\rm MP}$ / 4$_{\rm MP}$           & \textbf{71.0} & \textbf{31.6} & \textbf{0} \\
\bottomrule
\end{tabular}
\label{tab_resnet18_result}
\end{table}

\section{Experiments}
In this section, we conduct experiments on three lightweight models (\ie, ResNet18, MobileNetv2, and EfficientNetLite-B0) and three datasets (\ie, ImageNet, Pets, and CIFAR100). Please refer to the \textit{Supplementary Materials} for more detailed experimental setups. 

\subsection{ImageNet Classification}

\begin{table}[t] 
\setlength{\tabcolsep}{1.0mm}
\centering
\fontsize{8.3}{13.8}\selectfont
\caption{Accuracy and efficiency results for MobileNetV2.
$^{\dagger}$: QBitOPT uses channel-wise quantization to retain performance. 
}
\begin{tabular}{ccccc}
\toprule
Method   & \makecell{Bit-width \\(W/A)}     & \makecell{Top-1 Acc. \\(\%) $\uparrow$}     & \makecell{BitOPs \\(G) $\downarrow$} & \makecell{Retrain Cost \\(Epoch) $\downarrow$} \\ \midrule 
Baseline     & 32 / 32          &  72.6 &  -   & - \\
\hdashline
LSQ \cite{esser2020learned}     & 3 / 3          &  65.2     & 3.4    & 90 \\
QBR \cite{han2021improving} & 3 / 3        & \underline{67.4}  &  3.4   & 90 \\
HMQ \cite{habi2020hmq}      &    2$_{\rm MP}$ / 4$_{\rm MP}$     &   64.5   & -          & 50  \\
QBitOPT$^{\dagger}$ \cite{peters2023qbitopt} & 3$_{\rm MP}$ / 3$_{\rm MP}$ & 65.7 & - & \underline{30} \\
NIPQ \cite{shin2023nipq} & 3$_{\rm MP}$ / 3$_{\rm MP}$ & 62.3 & - & 43 \\
\rowcolor{yellow!25} Ours     & 3$_{\rm MP}$ / 3$_{\rm MP}$        & \textbf{67.8} & 3.6 & \textbf{0} \\ 
\hdashline
LSQ \cite{esser2020learned}     & 4 / 4          &  69.5  &  5.4    & 90 \\
EWGS \cite{lee2021network} &  4 / 4        & 70.3 &  5.4    & 100 \\
AdaBits  \cite{jin2020adabits}              & 4 / 4               & 70.4        & 5.4      &  \textbf{0} \\
QBR \cite{han2021improving} &  4 / 4        & \underline{70.4} &  5.4    & 90 \\
MPDNN  \cite{uhlich2019mixed}    & 3.75$_{\rm MP}$ / 4$_{\rm MP}$        & 69.8      & -         & 50 \\
QBitOPT$^{\dagger}$ \cite{peters2023qbitopt} & 4$_{\rm MP}$ / 4$_{\rm MP}$ & 69.7 & - & \underline{30} \\
NIPQ \cite{shin2023nipq}  & 4$_{\rm MP}$ / 4$_{\rm MP}$ & 69.2 & - & 43 \\
BayesianBits \cite{van2020bayesian} &4$_{\rm MP}$ / 4$_{\rm MP}$ & 69.0 & 5.9 & 40 \\
GMPQ \cite{han2021improving} & $\sim$ 4$_{\rm MP}$ / 4$_{\rm MP}$        & 70.4 & 7.4  & 40 \\
HAQ  \cite{wang2019haq}    & 6$_{\rm MP}$ / 4$_{\rm MP}$        &  69.5   &   8.3       & 30 \\
% \hdashline
\rowcolor{yellow!25} Ours     & 4$_{\rm MP}$ / 4$_{\rm MP}$        & \textbf{70.7} & 5.5 & \textbf{0}\\
\bottomrule
\end{tabular}
\label{tab_mobilenetv1_result}
\end{table}

\textbf{ResNet.} PACT demonstrates accuracy with 3-bits for both weights and activations, achieving 68.1\%. LSQ reaches 69.4\% accuracy but requires 90 retraining epochs. EdMIPS and GMPQ employ MPQ (3${\rm MP}$ / 3${\rm MP}$) for 68.2\% and 68.6\% accuracy but still require  considerable retraining costs. DNAS and FracBits adopt longer retraining epochs and yield better accuracy. 

When increasing the bit-width to 4-bits, PACT achieves 69.2\% accuracy with 35.0G BitOPs, while LSQ reaches 70.5\% accuracy. DNAS and FracBits demonstrate a 4-bits MPQ with slightly different results, while LIMPQ and SEAM both achieve the highest accuracy but still need 90 retraining epochs. 
Notably, our method with varying bit-width configurations (2${\rm MP}$/3${\rm MP}$, 3${\rm MP}$/3${\rm MP}$, and 4${\rm MP}$/4${\rm MP}$). The 4${\rm MP}$/4${\rm MP}$ configuration achieves the highest accuracy in the table at 71.0\%, with competitive BitOPs (31.6G) and no retraining cost. \\
\textbf{MobileNetV2.} 
QBR demonstrates a competitive Top-1 accuracy of 67.4\% with 3/3 bit-width and 3G BitOPs. QBitOPT adopts a performance-friendly channel-wise quantization and achieves 65.7\% accuracy in the 3${\rm MP}$/3${\rm MP}$ configuration and requires retraining split into 15 + 15 epochs \cite{peters2023qbitopt}, suggesting a more complex process. 
In the 4/4 bit-width category, QBR stands out with 70.4\% accuracy and 5.4G BitOPs, demonstrating efficiency. GMPQ delivers 70.4\% accuracy but requires 40 retraining epochs. HAQ achieves 69.5\% accuracy but incurs higher BitOPs (8.3G) and demands 30 retraining epochs. 

With 3$_{\rm MP}$/3$_{\rm MP}$ bit-width, our method reaches 67.8\% accuracy with 3.6G BitOPs and no retraining. Moreover, in the 4$_{\rm MP}$/4$_{\rm MP}$ configuration, it excels with a Top-1 accuracy of 70.7\% and competitive BitOPs (5.5G), all while eliminating retraining costs. \\
\begin{table}[t!]
\caption{Accuracy and efficiency results for EfficientNetLite-B0. $^{\dagger}$: QBitOPT uses channel-wise quantization to retain performance. 
}
\setlength{\tabcolsep}{1.0mm}
\centering
\fontsize{8.5}{13.8}\selectfont
\begin{tabular}{ccccc}
\toprule
Method     &  \makecell{Bit-width \\(W/A)}  & \makecell{Top-1 Acc.\\(\%) $\uparrow$}   & \makecell{BitOPs \\(G) $\downarrow$} & \makecell{Retrain Cost \\(Epoch) $\downarrow$} \\ \midrule
Baseline     & 32 / 32          &  75.4  &  -   & - \\
\hdashline
LSQ \cite{esser2020learned}  & 3 / 3      &  69.7  &  4.2     & 90 \\
NIPQ \cite{shin2023nipq}  & 3$_{\rm MP}$ / 3$_{\rm MP}$      & 66.5      & -      & 43 \\
QBitOPT$^{\dagger}$ \cite{peters2023qbitopt}  & 3$_{\rm MP}$ / 3$_{\rm MP}$      & \underline{70.0}      & -      & \underline{30} \\
MPQDNN \cite{uhlich2019mixed}  & 3$_{\rm MP}$ / 3$_{\rm MP}$      & 68.8      & -      & 50 \\
\rowcolor{yellow!25}Ours   & 3$_{\rm MP}$ / 3$_{\rm MP}$      & \textbf{70.4}  &    4.5   & \textbf{0} \\
\hdashline 
LSQ \cite{esser2020learned}  & 4 / 4      &  72.3  &  6.8  & 90 \\
NIPQ \cite{shin2023nipq}  & 4$_{\rm MP}$ / 4$_{\rm MP}$      & 72.3      & -      & 43 \\
QBitOPT$^{\dagger}$ \cite{peters2023qbitopt}  & 4$_{\rm MP}$ / 4$_{\rm MP}$      & \textbf{73.3}      & -      & \underline{30} \\
\rowcolor{yellow!25}Ours     & 4$_{\rm MP}$ / 4$_{\rm MP}$      & \underline{73.2}   & 6.9      & \textbf{0} \\
\bottomrule
\end{tabular}
\label{tab_efficientnet_result}
\end{table}

\noindent
\textbf{EfficientNet.} LSQ achieves 69.7\% accuracy with 4.2G BitOPs and requires 90 retraining epochs. In contrast, our 3${\rm MP}$/3${\rm MP}$ method attains 70.4\% accuracy with 4.5G BitOPs but eliminates the need for retraining, showcasing improved accuracy at a lower cost. 
QBitOPT achieves 70.0\% accuracy under 3${\rm MP}$/3${\rm MP}$ with 30 epochs for retraining. Our method at the same setting achieves 70.4\% accuracy without any retraining, highlighting superior performance without complex retraining. 
While LSQ and NIPQ achieve 72.3\% accuracy at 4/4 bit-width, they demand 90 retraining epochs. Our 4${\rm MP}$/4${\rm MP}$ method surpasses both, achieving 73.2\% accuracy with 6.9G BitOPs and no retraining. 
Our method consistently achieves comparable or superior accuracy with no retraining costs, demonstrating efficacy and simplicity in EfficientNet quantization. 

\begin{table}[t!]
\caption{Accuracy and efficiency results for ResNet with knowledge distillation. 
}
\setlength{\tabcolsep}{1.0mm}
\centering
\fontsize{8.5}{13.8}\selectfont
\begin{tabular}{ccccc}
\toprule
Method     &  \makecell{Bit-width \\(W/A)}  & \makecell{Top-1 Acc.\\(\%) $\uparrow$}   & \makecell{BitOPs \\(G) $\downarrow$} & \makecell{Retrain Cost \\(Epoch) $\downarrow$} \\ \midrule
Baseline     & 32 / 32          &  70.5  &  -   & - \\
\hdashline
GMPQ \cite{wang2021generalizable}  & 3$_{\rm MP}$ / 3$_{\rm MP}$      & 69.5      & 22.8      & 90 \\
SEAM \cite{tang2023seam}   & 3$_{\rm MP}$ / 3$_{\rm MP}$      & \underline{70.7}      & 23.0      & 90 \\
EQNet \cite{xu2023eq}   & 3$_{\rm MP}$ / 3$_{\rm MP}$      & 69.8      & -      & \textbf{0} \\
SDQ \cite{huang2022sdq}   &  3$_{\rm MP}$ / 3        & 70.2      & 25.1      & 90 \\
\rowcolor{yellow!25}Ours   & 3$_{\rm MP}$ / 3$_{\rm MP}$      & \textbf{70.9}      & 23.9      & \textbf{0} \\
\hdashline
NIPQ \cite{shin2023nipq} &   4$_{\rm MP}$ / 4$_{\rm MP}$        & 71.2      & 34.2      & 40 \\
SDQ \cite{huang2022sdq}   &  4$_{\rm MP}$ / 3        & \textbf{71.7}      & \underline{33.4}      & 90 \\
\rowcolor{yellow!25}Ours     & 4$_{\rm MP}$ / 4$_{\rm MP}$      & \underline{71.6}      & \textbf{31.6}      & \textbf{0} \\
\bottomrule
\end{tabular}
\label{tab_resnet18_distilled_result}
\end{table} 
\subsection{Ablation Study}
\textbf{Efficientness with KD.}
In comparison to the existing methods in Tab. \ref{tab_resnet18_distilled_result} when knowledge distillation (KD) is enabled with a ResNet101 teacher model, our method exhibits compelling advantages. 
GMPQ achieves a respectable 69.5\% accuracy with 3$_{\rm MP}$ bit-width but requires 90 retraining epochs. Our method surpasses it significantly, achieving a 70.9\% accuracy without retraining. Similarly, SEAM marginally improves accuracy to 70.7\%, but our method still outperforms with 70.9\% accuracy and no retraining costs. 
EQNet stands out with zero retraining epochs but falls significantly short of our method in accuracy (69.8\%). SDQ shows varied performance, but our method consistently outperforms it, particularly with 3$_{\rm MP}$ / 3$_{\rm MP}$ and 4$_{\rm MP}$ / 4$_{\rm MP}$ bit-width configurations, achieving higher accuracy and requiring no retraining compared to SDQ's 90 retraining epochs. \\

\begin{table}[t]
\centering
\fontsize{8.3}{13.8}\selectfont
\caption{Effectiveness of proposed dynamic bit-width schedule scheme and information distortion mitigation (IDM) training technique. To save costs, we train the weight-sharing model 80 epochs. }
\begin{tabular}{ccc}
\toprule 
Dynamic Bit Schedule       & IDM Training                     & 4 Bit Top-1 Acc. (\%) \\ \midrule
\xmark                         & \xmark                                & 68.3  \\
\cmark                          & \xmark                             &  69.1 \improvedperf{0.8}    \\
\cmark                          & \cmark                               & 69.5 \improvedperf{1.2}  \\ \bottomrule
\end{tabular}
\label{tab_ab_study}
\end{table}

\begin{figure}[t] 
\includegraphics[width=0.49\textwidth]{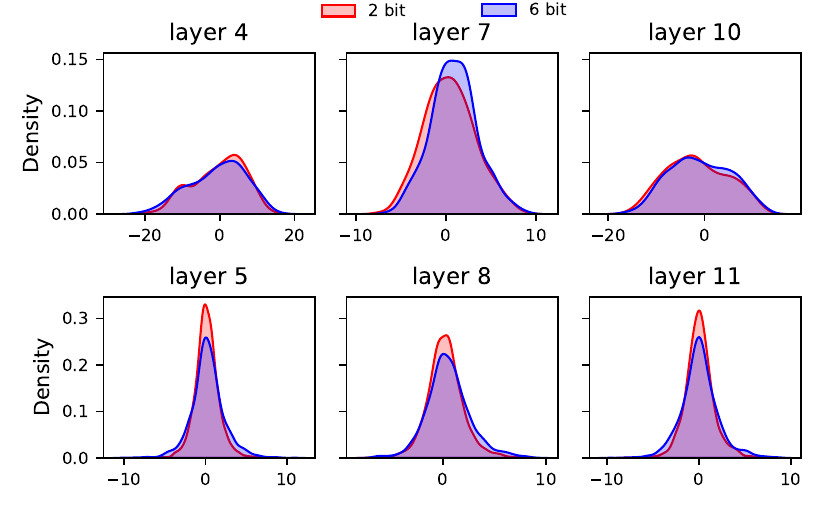}
\caption{Output density at 2bit and 6bits with our IDM training. Compared with \cref{fig:density_of_2_6bits}, information distortion of the small bit-widths is significantly mitigated. } 
\label{fig:density_of_2_6bits_ours}
\end{figure}

\noindent
\textbf{Effectiveness of Proposed Techniques.} 
\cref{tab_ab_study} investigates the impact of a dynamic bit-width schedule and our information distortion mitigation (IDM) training technique on the weight-sharing model. 
It presents three experimental scenarios: without both dynamic bit scheduling and IDM training resulting in 68.3\% Top-1 accuracy, dynamic bit scheduling alone with an improvement to 69.1\%, and the combination of both techniques achieving the highest Top-1 accuracy of 69.5\%. 
The results suggest that both dynamic bit scheduling and IDM training contribute positively to the model's performance, and their combination yields the most significant improvement. 
Moreover, our IDM training technique significantly mitigates  information distortion, as shown in \cref{fig:density_of_2_6bits_ours}. 

\subsection{Transfer Learning} 
We transfer the quantized weights for downstream benchmarks to verify the generalization ability of the proposed method. 
We directly use the pretrained checkpoints on ImageNet and then finetune the classifiers. 
As shown in \cref{tab:transfer_learning}, our method achieves the same accuracy as a full-precision model at 4-bits with smaller model complexity, which further confirms the superiority of the proposed method. 

\begin{table}[t]
\fontsize{8.5}{13.8}\selectfont 
\caption{Performance of transfer learning using the pretrained weights on ImageNet. } 
\begin{tabular}{cccc} 
\toprule
Model                        & \makecell{Bit-width \\ (W/A)} & \makecell{CIFAR100 \cite{krizhevsky2009learning} \\ Top-1 Acc. (\%)} & \makecell{Pets \cite{parkhi2012cats} \\ Top-1 Acc. (\%)} \\ \midrule
\multirow{3}{*}{ResNet18}    & 32 / 32                       & 79.4         &  88.9    \\
                             & 4$_{\rm MP}$ / 4$_{\rm MP}$   & 79.5 \improvedperf{0.1}        &   88.7 \decreasedperf{0.2}   \\
                             & 3$_{\rm MP}$ / 3$_{\rm MP}$   & 78.7   \decreasedperf{0.7}       & 87.9  \decreasedperf{2.0} \\ \midrule
\multirow{3}{*}{MobileNetV2} & 32 / 32                       & 78.9         &   86.0   \\
                             & 4$_{\rm MP}$ / 4$_{\rm MP}$   & 79.0 \improvedperf{0.1}        & 86.1 \improvedperf{0.1} \\
                             & 3$_{\rm MP}$ / 3$_{\rm MP}$   & 78.2   \decreasedperf{1.7}       &  84.1  \decreasedperf{1.9} \\ \bottomrule
\end{tabular}
\label{tab:transfer_learning}
\end{table}
\section{Conclusion}
In this paper, we introduce a novel one-shot training-searching paradigm for mixed-precision model compression. More specifically, traditional approaches focus on bit-width configurations but overlook significant retraining costs. We identified and addressed bit-width interference issues by introducing a dynamic scheduler and an information distortion mitigation technique. 
Together with an inference-only greedy search scheme, our method can significantly reduce the costs of mixed-precision quantization. Experiments on three commonly used benchmarks across various network architectures validate the effectiveness and efficiency of the proposed method in compressing models. Overall, our method offers a promising solution for deploying compressed models without compromising performance on resource-limited devices. 

\section*{Acknowledgment}
This work was supported by the National Key Research and Development Program of China~No.~2023YFF0905502, National Natural Science Foundation of China (Grant No.~62250008), Beijing National Research Center for Information Science and Technology (BNRist) under Grant No.~BNR2023TD03006 and Beijing Key Lab of Networked Multimedia, and Shenzhen Science and Technology Program (Grant No.~RCYX20200714114523079 and No.~JCYJ20220818101014030). 

{
    \small
    \bibliographystyle{ieeenat_fullname}
    \bibliography{main}
} 

% \input{appendix}


\end{document}